\documentclass{article}
\usepackage[numbers]{natbib}
\usepackage{arxiv}

\usepackage[utf8]{inputenc} 
\usepackage[T1]{fontenc}    
\usepackage{url}            
\usepackage{booktabs}       
\usepackage{amsfonts}       
\usepackage{nicefrac}       
\usepackage{microtype}      

\usepackage{lipsum}         
\usepackage{graphicx}
\usepackage{natbib}
\usepackage{doi}

\usepackage{xcolor} 
\usepackage[ruled]{algorithm2e}

\usepackage{amsmath, float}
\usepackage{hyperref}
\usepackage{cleveref}       
\usepackage{blkarray}
\usepackage{subcaption}	
\usepackage{caption}
\usepackage{multirow}

\newcommand{\R}{\mathbb{R}}
\newcommand{\nys}{\textnormal{nys}}
\newcommand{\diag}{\textnormal{diag}}
\newcommand{\bnt}{\textnormal{bn}}

\def\pp{\textcolor{black}}

\def\zl{\textcolor{black}}
\def\hz{\textcolor{black}}
\def\kh{\textcolor{black}}
\def\meka{\textcolor{black}}

\title{Boosting Nystr\"{o}m Method}

\date{}

\author{ 
Keaton Hamm\thanks{These two authors contributed equally to this work.}\\
    Department of Mathematics\\
    University of Texas at Arlington\\
    Arlington, TX 76019, USA\\
	\texttt{keaton.hamm@uta.edu} \\
	\And
Zhaoying Lu\footnotemark[1]\\
	Department of Mathematics\\
	University of Arizona\\
	Tucson, AZ 85719, USA \\
	\texttt{zhaoyinglu@arizona.edu} \\
	\AND
Wenbo Ouyang\\
	Department of Mathematics\\
	University of Arizona\\
	Tucson, AZ 85719, USA \\
	\texttt{wenboouyang@arizona.edu} \\
\And
Hao Helen Zhang\thanks{corresponding author}\\
   Department of Mathematics \\
   University of Arizona \\
   Tucson, AZ 85721, USA \\
   \texttt{hzhang@math.arizona.edu}
   }

\renewcommand{\shorttitle}{Boosting Nystr\"{o}m Method}

\hypersetup{
pdftitle={A template for the arxiv style},
pdfsubject={q-bio.NC, q-bio.QM},
pdfauthor={David S.~Hippocampus, Elias D.~Striatum},
pdfkeywords={First keyword, Second keyword, More},
}

\begin{document}
\maketitle

\begin{abstract}
The Nystr\"{o}m method is an effective tool to generate low-rank approximations of large matrices, and \hz{it} is particularly useful for kernel-based learning. To improve the standard Nystr\"{o}m approximation, ensemble Nystr\"{o}m algorithms compute a mixture of Nystr\"{o}m approximations which are generated independently based on column resampling. We propose a new family of algorithms, {\it boosting Nystr\"{o}m}, which iteratively generate multiple ``weak'' Nystr\"{o}m approximations (each using a small number of columns) in a sequence adaptively - each approximation aims to compensate for the weaknesses of its predecessor - and then combine them to form one strong approximation. We demonstrate that our boosting Nystr\"{o}m algorithms can yield more efficient and accurate low-rank approximations to kernel matrices. Improvements over the standard and ensemble Nystr\"{o}m methods are illustrated by simulation studies and real-world data analysis.
\end{abstract}


\section{Introduction}
Kernel-based learning is widely used to identify complex nonlinear patterns and \hz{variable} relationships in massive and high-dimensional data.
Well-known kernel learning methods include support vector machines (SVMs)~\cite{cortes1995support}, kernel principle component analysis (PCA)~\cite{scholkopf1998nonlinear}, ANOVA kernel ridge regression~\cite{saunders1998ridge}, spectral clustering \cite{ng2001spectral,shi2000normalized}, and Gaussian process classification~\cite{williams1998bayesian}.
One major challenge in implementing 
kernel methods is that evaluation of \hz{the} entire kernel matrix, or the Gram matrix $G\in\R^{n\times n}$, can be costly when the sample size $n$ is enormously large. In addition, the dense structure of a kernel matrix precludes use of sparsity-based techniques for speed up. 

\pp{In many cases}, kernel matrices are incoherent and approximately low-rank \cite{talwalkar2010matrix}, for example, the datasets of MNIST~\cite{lecun1998mnist} and PIE~\cite{sim2002cmu}, making it possible to utilize their low-rank approximation to reduce computation cost in optimization and implementation. The Nystr\"{o}m method~\cite{578947ad5de84c50b06a6bab4a2cfc40} is an effective tool to generate low-rank approximations for kernel matrices, \hz{and has shown successful performance} in diverse areas such as image processing, natural language processing, and bioinformatics~\cite{fowlkes2004spectral,kumar2009ensemble, talwalkar2008large, 578947ad5de84c50b06a6bab4a2cfc40, zhang2008improved}. Other recent related works on kernel approximation and Nystr\"{o}m approximation \hz{include} Nystr\"{o}m kernel mean embeddings~\cite{chatalic2022nystr}, random features~\cite{Liu2021RandomFF}, and CUR decomposition~\cite{cai2021robust}.
The \pp{standard} Nystr\"{o}m method \hz{is} a numerical quadrature method \pp{initially} designed to solve integral equations~\cite{baker1977numerical}. Over the years, its use has been adapted to the machine learning context to construct low-rank approximations to symmetric positive semidefinite (SPSD) Gram matrices $G$ based on sampling only a small number, $m\ll n$, of columns of $G\in\R^{n\times n}$~\cite{drineas2005nystrom,gittens2016revisiting,578947ad5de84c50b06a6bab4a2cfc40}. Many extensions of the Nystr\"{o}m method have been designed to improve approximation accuracy (typically $\|G-CW_k^\dagger C^{\top}\|_F$, where $\|\cdot\|_F$ is the Frobenius norm), including density-weighted Nystr\"{o}m~\cite{zhang2009density}, modified Nystr\"{o}m~\cite{wang2013improving}, SS-Nystr\"{o}m~\cite{wang2014modified} (an extension to the former), and ensemble Nystr\"{o}m~\cite{kumar2009ensemble}. Other column sampling schemes such as the sparse greedy matrix approximation~\cite{smola2000sparse}, $k$-means-based sampling~\cite{zhang2008improved}, and adaptive sampling techniques~\cite{deshpande2006matrix, kumar2012sampling} have been proposed \pp{to improve the uniform sampling technique of the} standard Nystr\"{o}m approximation. These works demonstrate that low-rank approximations of kernel matrices can facilitate kernel learning with improved computational efficiency while maintaining accuracy.

\subsection{Motivations and contributions of this work}

The standard Nystr\"{o}m method is successful in large-scale applications as it \hz{can substantially reduce} the cost of storage and computation by constructing low-rank approximations of the kernel matrix with \hz{only} a small number of columns from the kernel matrix. \pp{In recent years, various efforts have been made to improve the original Nystr\"{o}m method}, \hz{and they} are highlighted as follows.  
Firstly, one can use alternative sampling schemes, rather than the simple uniform sampling employed in the standard Nystr\"{o}m method, to sample more representative columns. Column sampling plays a critical role in both theory and empirical performance of the method. \meka{Some complex data-dependent sampling techniques have been introduced, such as  Determinantal Point Processes~\cite{li2016fast} and Ridge Leverage Scores~\cite{musco2017recursive, rudi2018fast}, to improve the quality of the low-rank approximation}. They \hz{are shown to achieve superior performance but at the expense of much higher} computational cost compared with uniform sampling. Adaptive sampling techniques~\cite{deshpande2006matrix, kumar2012sampling,musco2017recursive} are proposed to sample columns iteratively, based on the probability distribution proportional to the reconstruction error, and shown to select more representative columns. Secondly, \pp{instead of sampling from the original columns, some find that it would be more effective to first form columns into clusters and then sample from the cluster centers}~\cite{zhang2008improved}. For example, $k$-means clustering has been used~\cite{kumar2012sampling}.
Thirdly, the ensemble Nystr\"{o}m approximation~\cite{kumar2009ensemble} has been proposed to \pp{generate} multiple Nystr\"{o}m approximations and combine them as a weighted mixture, which \hz{can} provide a more accurate approximation. The mixture weights \hz{are} chosen as uniform weights, exponential weights, or computed from a ridge regression. \meka{Lastly, the recently proposed method Memory Efficient Kernel Approximation (MEKA)~\cite{si2014memory}
} \hz{applies the block-cluster structure and low-rank techniques to shift-invariant kernels and achieves effective approximation accuracy. Their method is specifically designed for shift-invariant kernels rather than general kernels}. 

Among the aforementioned methods, ensemble Nystr\"{o}m provides \meka{one of} the best low-rank approximation\zl{s} of the kernel matrix in terms of accuracy. However, ensemble methods generate multiple standard Nystr\"{o}m approximations independently in parallel, which requires training a large number of learners to get good performance and share similar limitations as bagging~\cite{breiman1996bagging}. In this paper, we propose the idea of {\it boosting Nystr\"{o}m}, which creates multiple weak learners (here approximations) sequentially and adaptively to improve the overall approximation. For each approximation, its column sampling is guided by the result of previous approximations, trying to remedy the weaknesses of its forerunners and \hz{gradually} build a strong approximation model. 

Boosting is an \pp{ensemble learning technique that constructs weak learners iteratively and combines them to form a strong learner with improved prediction accuracy~\cite{freund1995boosting, freund1997decision,schapire1990strength}. Inspired by the success of boosting, gradient boosting~\cite{friedman2001greedy} and Xgboost~\cite{chen2016xgboost}, we develop a general class of boosting algorithms, using the Nystr\"{o}m approximation as a base learner, to build a more accurate low-rank approximation for the kernel matrix. Each approximation is a weak learner as only a small number of columns is sampled. Compared to ensemble and adaptive Nystr\"{o}m approximation, boosting Nystr\"{o}m has several major differences and advantages: (i) Instead of training a large number of independent learners as done by ensemble Nystr\"{o}m, the boosting algorithm trains a much smaller number of weak learners sequentially to build a strong learner, which can substantially speed up the computation; (ii) During training, the boosting algorithm keeps track of the overall approximation error and adaptively adjusts the weights of different approximations, where a better approximation receives a higher weight, which can lead to significant accuracy improvement over standard and ensemble Nystr\"{o}m}; \pp{(iii) The proposed boosting algorithm has a general and flexible framework, \hz{which allows} users to explore a variety of sampling schemes (uniform, exponential, cluster-based) and model averaging methods (equal weights, exponential weights, or weights estimated from ridge regression), and to identify the optimal configuration for a given kernel matrix. }
\pp{Furthermore, our numerical} experiments \pp{suggest} that boosting Nystr\"{o}m is capable of achieving consistently better accuracy with fewer Nystr\"{o}m approximations than the ensemble and adaptive Nystr\"{o}m methods.

\subsection{Organization}

The remainder of this paper is organized as follows. In Section \ref{SEC:Alg}, we \pp{introduce}
notation \pp{used in this paper} and give a brief overview of the \pp{standard} Nystr\"{o}m and ensemble Nystr\"{o}m methods. Then we propose boosting Nystr\"{o}m, present the algorithm and discusss \kh{several} variants \pp{and their practical implementation}. Section \ref{SEC:Experiments} presents the experimental results on simulated and real data. We compare the performance of our boosting Nystr\"{o}m methods with ensemble Nystr\"{o}m \pp{in terms of their approximation accuracy for reconstructing} the kernel matrix and the computation time.

\section{Boosting Nystr\"{o}m}\label{SEC:Alg}
\subsection{Notation}
\pp{Throughout the paper, we use} $G\in\R^{n\times n}$ \pp{to} denote a symmetric, positive semidefinite (SPSD) kernel matrix, whose Singular Value Decomposition (SVD) takes the form $G=U\Sigma U^{\top}$\pp{,} where $U$ \pp{is} an orthogonal matrix and $\Sigma=\diag(\sigma_1,\dots,\sigma_n)$ contains \pp{ordered} singular values $\sigma_1\geq\dots\geq\sigma_n\geq0.$  Given an index set $I=\{i_1,\dots,i_m\}\subset\{1,\dots,n\}$, \pp{let} $C=G[:,I]$ denote the submatrix of $G$ formed \pp{by} the columns indexed by $I$, and \pp{let} $W=G[I,I]$ \pp{be} the $m\times m$ submatrix of $G$ formed by the intersection of $C$ and $C^{\top}$. Given a symmetric matrix $A = U\Sigma U^{\top} \in\R^{n\times n}$, \pp{let} $A_k=U_k\Sigma_k U_k^{\top}$ \pp{denote} the best rank-$k$ approximation to $A$ (in the Frobenius norm), where $U_k$ is the $n\times k$ matrix consisting of the first $k$ columns of $U$, and $\Sigma_k=\diag(\sigma_1,\dots,\sigma_k).$  The Moore--Penrose pseudoinverse of $A_k$ is \pp{denoted by} $A_k^\dagger = U_k\Sigma_k^\dagger U_k^{\top}$, where $\Sigma_k^\dagger = \diag(\frac{1}{\sigma_1},\dots,\frac{1}{\sigma_k}).$  The Frobenius norm of a matrix is given by $\|A\|_F = (\sum_{i,j}|A_{i,j}|^2)^\frac12.$

\pp{For the proposed} boosting Nystr\"{o}m algorithm, we \pp{need to} select validation sets \pp{during the training process}. If $V\subset\{1,\dots,n\}\times\{1,\dots,n\}$, then we use $G^V$ to denote the submatrix of $G$ whose indices correspond to $V$, i.e., if $V = I\times I$ for some $I$ as above, then $G^V=G[I,I].$  \pp{Furthermore, we use} $p$ to denote the total number of learners used in either the boosting or \hz{the} ensemble Nystr\"{o}m.

\subsection{Review of Standard Nystr\"{o}m}

The Nystr\"{o}m method is a \pp{numerical tool for computing} solutions to integral equations \cite{baker1977numerical}. Williams and Seeger~\cite{578947ad5de84c50b06a6bab4a2cfc40} show that it can be used to find approximations to SPSD kernel matrices by approximating their eigenvectors and eigenvalues. For kernel learning with large-scale data, the Nystr\"{o}m method is \hz{particularly} useful to generate a low-rank approximation of the kernel matrix using only a subset of its columns. 

\pp{Given a positive integer $k$, the} standard Nystr\"{o}m method provides a rank-$k$ approximation of $G$ \pp{of} the form $G^{\nys} = CW_{k}^\dagger C^{T}$, where $C=G[:,I]$ for some $I$ of size $m\ll n$ (recall that $W_k$ is the best rank-$k$ approximation to the small submatrix of $G$). More specifically, suppose that $I = \{1,\dots,m\}$, then we can write
\[G = \begin{bmatrix} W & B^{\top}\\ B & E\end{bmatrix}, \quad C = \begin{bmatrix} W\\B\end{bmatrix},\]
and $G^{\nys} = CW_k^\dagger C^{\top}$. Tropp et al.~\cite{tropp2017fixed} (see also Pourkamali-Anaraki and Becker \cite{pourkamali2019improved}) show that a Nystr\"{o}m approximation of the form $(CW^\dagger C^{\top})_k$ provides an accurate low-rank approximation of kernel matrices in many instances. The computational complexity of the standard Nystr\"{o}m is $O(m^3+nmk)$.

\subsection{Review of Ensemble Nystr\"{o}m}

\pp{The ensemble Nystr\"{o}m method \cite{kumar2009ensemble} first builds $p$ independent learners, where each learner is constructed using a subset of $m$
columns, and then combines them using a weighted average. In particular, the ensemble method} samples $mp$ columns of $G$ uniformly and splits the selected columns into $p$ \pp{sets} 
of $m$ columns, using each \pp{set} 
to generate a Nystr\"{o}m approximation $G^{\nys}_i$, $i \in \{1, \dots, p\}$ \pp{independently in parallel}. Then it combines these Nystr\"{o}m approximations $G^{\nys}_i$ with mixture weights $\mu_i$ to form an ensemble Nystr\"{o}m approximation $G^{\textnormal{ens}}=\sum_{i=1}^p\mu_iG^{\nys}_i$. The weights $\mu_i$ may be chosen in various ways; Kumar et al.~\cite{kumar2009ensemble} consider uniform weights, exponential weights, and ridge regression weights, which will be explained in more detail in the following subsection. The ensemble Nystr\"{o}m method also samples additional columns for tuning parameters in the exponential and ridge regression weight settings.

\subsection{Proposed Method: Boosting Nystr\"{o}m}

We propose a new class of boosting algorithms, called {\it Boosting Nystr\"{o}m}, to improve the standard Nystr\"{o}m and ensemble Nystr\"{o}m approximation from two perspectives: prediction accuracy and computational cost. The standard Nystr\"{o}m method only builds one single low-rank approximation, which may have issues of over-fitting, numerical instability, \hz{and} high variance. Though ensemble Nystr\"{o}m is designed to solve these issues by generating multiple approximations, it has two limitations: it requires training a large number of learners, which may increase the computational cost substantially; these multiple approximations are learned independently without sharing information, which may deliver suboptimal performance in terms of prediction accuracy. The proposed boosting Nystr\"{o}m \zl{method} aims to reduce the computation cost of ensemble methods \hz{and produce} a more accurate low-rank approximation of the kernel matrix. 

\pp{The key idea of boosting Nystr\"{o}m is as follows: we use the standard Nystr\"{o}m as a base learner to build multiple ``weak'' low-rank approximations in sequence, where each successor learner tries to tackle weaknesses of the previous learners. Then these learners are combined to construct a ``strong'' learner (an approximation to the kernel matrix). The learning procedure involves two main steps: boosting and strong approximation, as} summarized below.

\begin{itemize}
    \item \textbf{Initialization: Standard Nystr\"{o}m.} We start by sampling $m$ columns from $G$ uniformly to form the first standard Nystr\"{o}m approximation $G^{\nys}_1$.
    \item \textbf{Boosting (iterative part)}
    \begin{itemize}
        \item \textbf{Intermediate ensemble Nystr\"{o}m approximation.} In each iteration, we combine all the ``weak'' learners obtained so far with mixture weights $\mu_k$ \pp{to get} $\overline{G}^{\nys}_i = \sum_{k=1}^{i}\mu_kG^{\nys}_k$.
        \pp{Choices of mixture weights, are discussed in detail below}.
        
        \item \textbf{Sampling step.} In each iteration, a clustering algorithm is applied to the submatrix according to the validation set $V$ of the residual \pp{matrix}, $G^{V} - \overline{G}^{\nys, V}_{i}$, formed from the intermediate ensemble Nystr\"{o}m approximation  $\overline{G}^{\nys}_i$. \pp{The residual matrix is updated at each iteration and used to train the next learner}. The centers resulting from the clustering method are used as the new $m$ columns for generating a ``weak'' learner. \pp{Choices of clustering methods are discussed in the following}. 
        
        \item \textbf{``Weak'' Approximation.} The standard Nystr\"{o}m method is the ``weak'' or ``base'' learning algorithm and uses the $m$ columns obtained in the sampling step to generate a ``weak'' approximation $G^{\nys}_i$.
    \end{itemize}
    \item \textbf{Strong Approximation.} After $p$ iterations we obtain $p$ ``weak'' Nystr\"{o}m approximations. \hz{At} this step, we combine them with the mixture weights $w_i$ defined by the weighting method (see below), then generate the boosting Nystr\"{o}m approximation $G^{\textnormal{bn}} = \sum_{i=1}^{p}w_i G^{\nys}_i$.
\end{itemize}

\kh{Note that more sophisticated column sampling procedures could easily be employed in the} \meka{initialization step} \kh{of our algorithm. However, we find that the most na\"{i}ve of all sampling methods (uniformly sampling columns) performs very well when coupled with boosting. Therefore, we focus here on boosting and do not require use of more costly sampling schemes.}

\medskip
{\bf Choice of Mixture Weights}. \pp{The mixture weights $\mu_k$ and $w_i$ play a critical role at both} boosting and strong approximation step\pp{s}. \pp{During the learning process, we compute and update the weights (except in the uniform case) adaptively based on} the error of the Nystr\"{o}m approximation on a fixed randomly sampled validation set $V_1$ with tuned parameters. Define $\mathbf{w}=(w_1, \ldots, w_p)$. The following are three choices \hz{of} weights implemented in the algorithm. 

\begin{itemize}
  \item{\pp{Uniform Weights}.} Set all the weights as $w_i = \frac{1}{p}$, $i = 1,$ \dots, $p$, and similarly for $\mu_k$. This choice treats all learners equally regardless of their performance. Intuitively, learners giving better approximation should \hz{be assigned} higher weights, which motivates other weight choices \hz{described below}.
  
  \item{Exponential weights.} The weights are defined by $w_i = \exp(-\eta \hat{\epsilon_i})/Z$, where $\eta > 0$ is a fixed parameter, $Z$ is a normalization factor, and $\hat{\epsilon_i} = \|G^{\nys, V_1}_i - G^{V_1}\|$ represents the reconstruction error of $G^{\nys}_i$ over the validation set $V_1$. Here $\eta$ is a tuning parameter which \hz{is} selected based on the validation error. More details on parameter tuning will be provided later.
  
 \item{Weights estimated from ridge regression}. The weights can also be dynamically computed by fitting a ridge regression model \hz{which minimizes} the objective function $f(\pp{\mathbf{w}}) = ||\sum_{i=1}^{p}w_i G^{\nys, V_1}_i - G^{V_1}||^2_F + \lambda ||\textbf{w}||^2_2$, where $\lambda$ is a tuning parameter. One main advantage of using ridge regression is \hz{its} closed-form solution, which can greatly facilitate computation.
\end{itemize}

\medskip
\noindent{\bf{Choices of Tuning Parameters}}. Our boosting Nystr\"{o}m procedure involves a few tuning parameters, including $\eta$ for exponential weights, $\lambda$ in ridge regression, and the number of columns $m$ \kh{in} the sampling step. Their choices are critical to the performance of the final approximation. To find optimal values of $\eta$ and $\lambda$, we randomly sample a validation set $V_2$ at the beginning of our algorithm and determine their values based on the tuning error. Since we tune these parameters \hz{at} the intermediate ensemble Nystr\"{o}m approximation \hz{of} each iteration, the strong approximation step can be skipped if we use the same weighting methods in the boosting and strong approximation steps, i.e., $G^{\textnormal{bn}} = \overline{G}^{\nys}_p$.

\medskip
\noindent{\bf{Choices of Clustering Methods.}} At the sampling step \hz{of} each iteration, we uniformly sample the validation set $V$ from $G$, but exclude the columns in the validation sets $V_1$ and $V_2$ and those used for generating previous ``weak'' learners. It's worth noting that the validation set $V$ in the sampling step is re-sampled at every iteration, while the others are not. \kh{We then cluster the columns of the residual $G^{V} - \overline{G}^{\nys, V}$ and use the nearest neighbors of the cluster centers to form the next ``weak" learner.} 
In the algorithm, we implement three types of clustering methods \hz{described} as follows. Their performance will be evaluated and compared in the numerical experiments.

\begin{itemize}
    \item \textbf{$k$-means.} $k$-means clustering~\cite{macqueen1967some} divides $n$ observations into \hz{$k$} clusters, with each observation assigned to the cluster with the closest mean. We cluster all the columns in the submatrix of the residual into $m$ groups, then \hz{use those columns which are closest to $k$  centers to generate new  ``weak'' learners}. 

    \item \textbf{$k$-medoids.} In contrast to $k$-means, \hz{the $k$-medoids method} employs medoids as cluster centers, so it uses actual data points as centers. The former is faster, but the latter is less sensitive to outliers. We use the Partitioning Around Medoids (PAM) algorithm~\cite{kaufman2009finding} in R to \pp{implement} the $k$-medoids \pp{clustering}.

    \item \textbf{$k$-medoids-SF.} To mitigate the computational \hz{cost of} $k$-medoids, we \hz{also consider} the simple and fast algorithm for $k$-medoids clustering ($k$-medoids-SF) proposed by Park and Jun~\cite{park2009simple}.

\end{itemize}


\begin{algorithm}[ht]
\SetKwInOut{Input}{Input}
\SetKwInOut{Output}{Output}
\Input{\pp{An} $n\times n$ SPSD Gram matrix $G$, \pp{the} number of Nystr\"{o}m approximations $p$, \pp{the} number of columns for each Nystr\"{o}m approximation $m$, \pp{the} number of columns $s$, $v_1$, $v_2$ in the validation sets $V$, $V_1$, and $V_2$ respectively, \pp{and the} target rank $k$.}
\Output{$G^{\bnt}$, the boosting Nystr\"{o}m approximation of $G$}

\textbf{Step 1. Standard Nystr\"{o}m.} Sample $m$ columns and generate the standard Nystr\"{o}m approximation $G^{\nys}_1 = CW_{k}^\dagger C^{T}$. Randomly select $v_1 + v_2$ columns as the validation sets $V_1$ and $V_2$.

\textbf{Step 2. Boosting.}

\For{$i=1$ to $p$}{

(a) compute the mixture weights $\mu_k$ and combine all $G^{\nys}_i$ to obtain $\overline{G}^{\nys}_i = \sum_{k=1}^{i}\mu_kG^{\nys}_k$; 

(b) resample $s$ columns from $G$ for the validation set $V$, excluding columns \hz{used by} previous ``weak'' learners and in the validation sets $V_1$ and $V_2$;

(c) compute residuals on $V$: $e_i$ = $G^{V} - \overline{G}^{\nys, V}_{i}$;

(d) cluster the columns in $e_i$ into $m$ groups and select $m$ new columns using the centers \hz{resulted} from clustering algorithm;

(e) compute the ``weak'' Nystr\"{o}m approximation $G^{\nys}_{i+1}= CW_{k}^\dagger C^{T}$};

\textbf{Step 3. Strong approximation.} Combine the $p$ ``weak'' Nystr\"{o}m approximations $G^{\nys}_i$ with mixtures weights $w_i$ and obtain \pp{the final approximation} $G^{\bnt} = \sum_{i=1}^{p}w_i G^{\nys}_i$.

 \Return{$G^{\bnt}$}
 \caption{Boosting Nystr\"{o}m method}\label{ALG:SC}
\end{algorithm}

We note that, regardless of the mixture weights, if Lloyd's algorithm \hz{is used} for $k$-means clustering then the total complexity of Algorithm \ref{ALG:SC} is $O(pn^2m)$, where the dominating complexity is the standard Nystr\"{o}m approximation in Step (e) of the for loop.

\section{Numerical experiments}\label{SEC:Experiments}

In this section, we present empirical assessment \hz{and comparison} of the ensemble Nystr\"{o}m method and \pp{the proposed} boosting Nystr\"{o}m \zl{method}. \hz{In our experiment, we implement the boosting Nystr\"{o}m using a variety of configurations listed} in Table \ref{TAB:Boosting}, \hz{including different ways of computing} mixture weights in the intermediate ensemble Nystr\"{o}m approximation and strong approximation steps, \hz{and three choices} of clustering algorithms \hz{at} step (d) of the Boosting part. For clarity, \hz{our figures illustrate the performance of a subset of the configurations}. 


Table \ref{TAB:Boosting} shows the naming scheme for variants of boosting Nystr\"{o}m based on the mixture weight choices in the boosting and strong approximation steps as well as the choice of clustering algorithm. The naming convention is XYB-clustering, where X = U, E, or R (Uniform, Exponential, or Ridge), \hz{which denotes} the choice of mixture weights \hz{at} the intermediate ensemble Nystr\"{o}m step in the Boosting part, Y = U, E, or R (Uniform, Exponential, or Ridge), \hz{representing} the choice of mixture weights in the strong approximation step, B stands for Boosting, and clustering = mean, med, or SF for $k$-means, $k$-medoids, and SF, respectively (SF is the simple and fast algorithm of \cite{park2009simple}). For example, if the weighting methods are uniform and ridge regression \hz{at} the intermediate ensemble Nystr\"{o}m approximation step and strong approximation step, respectively, and the clustering method is \emph{k}-means clustering \hz{at} the sampling step, the boosting Nystr\"{o}m algorithm is denoted \hz{as} URB-mean.

\begin{table*}[h]
    \centering
    \caption{Boosting Nystr\"{o}m methods}\label{TAB:Boosting}
    \resizebox{\textwidth}{!}{\begin{tabular}{lllllllllll}
        \toprule
        \multirow{2}{*}[-1em]{Clustering Methods} &
          {Boosting} & \multicolumn{3}{c}{Uniform} & \multicolumn{3}{c}{Exponential} & \multicolumn{3}{c}{Ridge}\\
        \addlinespace[10pt]
        \cmidrule(lr){2-2} \cmidrule(lr){3-5} \cmidrule(lr){6-8} \cmidrule(lr){9-11}\\
        {} & Strong Approximation & Uniform & Exponential & Ridge & Uniform & Exponential & Ridge & Uniform & Exponential & Ridge \\
        \midrule
        \emph{k}-means & {} & UUB-mean  & UEB-mean & URB-mean  & EUB-mean  & EEB-mean & ERB-mean & RUB-mean  & REB-mean & RRB-mean\\
        \emph{k}-medoids & {} & UUB-med & UEB-med & URB-med & EUB-med & EEB-med & ERB-med & RUB-med & REB-med & RRB-med\\
        \emph{k}-medoids-SF & {} & UUB-SF & UEB-SF & URB-SF & EUB-SF & EEB-SF & ERB-SF & RUB-SF & REB-SF & RRB-SF\\
        \bottomrule
    \end{tabular}}
\end{table*}

Since the performance of the algorithms may vary \hz{for different} kernel functions and data structures, we \hz{conduct} \pp{numerical experiments} on several simulated and real data sets using linear and Gaussian kernels to \hz{evaluate} the algorithms in different cases. \pp{As shown in the numerical results below, our} algorithms achieve significant performance gains in most situations \kh{as measured by a 2-sample $t$-test comparing the boosting method with ensemble Nystr\"{o}m}. The accuracy is measured by computing the relative error in the Frobenius norm, $\frac{||G^{\bnt} - G||_F}{||G||_F}$.

\subsection{Simulation}

For our synthetic experiments, we \hz{randomly sample} 1000 \hz{data points} from the standard normal distribution on $\R^2$ and \hz{generate} a kernel matrix $G$ via a Gaussian kernel with $\sigma = 1$. For each learner, we choose $m=10$ columns and set the target rank to be $k=10$. The number of columns in the validation sets $V$, $V_1$, and $V_2$ are set to 100, 20, and 20, respectively. For simplicity, we fix the parameter $\eta$ to be 0.01 in the exponential weight method in the following experiments, so $V_2$ is only used for tuning the parameter $\lambda$ in the ridge regression weight method.

\begin{figure}[h!]
  \centering
  \begin{subfigure}[b]{0.49\linewidth}
    \includegraphics[width=\linewidth]{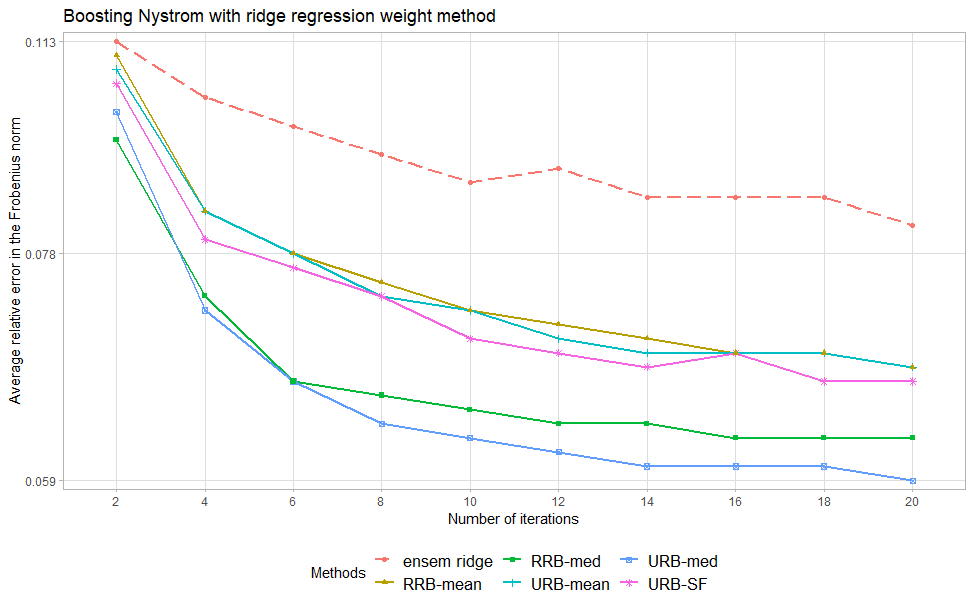}
    \caption{Boosting Nystr\"{o}m with ridge regression weight.}\label{fig:sima}
  \end{subfigure}
  \begin{subfigure}[b]{0.49\linewidth}
    \includegraphics[width=\linewidth]{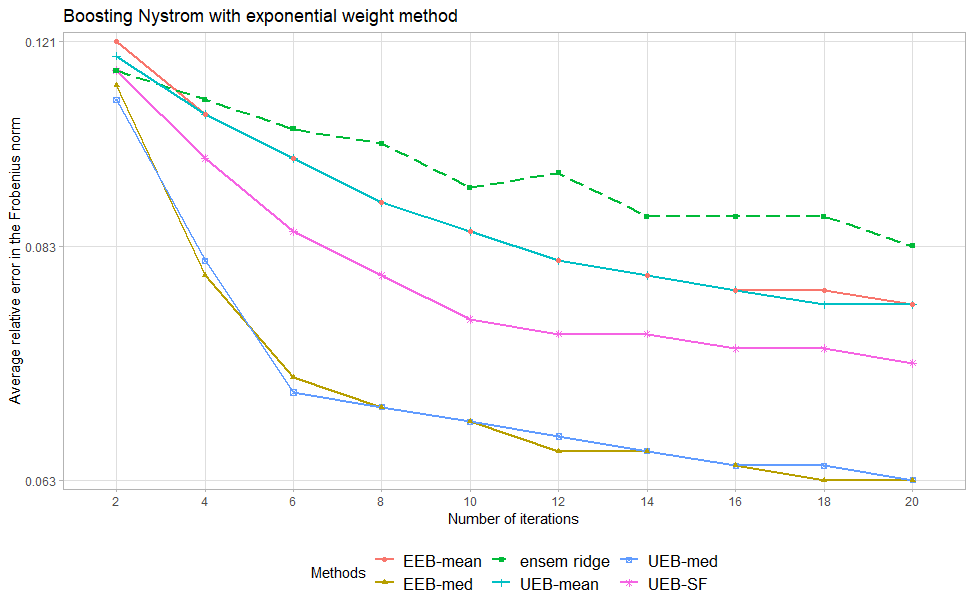}
    \caption{Boosting Nystr\"{o}m with exponential weight method.}\label{fig:simb}
  \end{subfigure}
  \caption{Average relative errors for 100 replicates of each random algorithm on a Gaussian kernel matrix formed from standard normal simulation data.}
  \label{fig:sim}
\end{figure}

We first illustrate the performance of boosting Nystr\"{o}m with ridge regression as the weight method of strong approximation. \hz{Its} average relative errors in the Frobenius norm of 100 replication studies are presented in Figure \ref{fig:sima}. The results of ensemble Nystr\"{o}m method (dashed line) \pp{are} also included in the graph \hz{for} comparison. \hz{In the figure, the $x$-axis represents} the number of iterations for the boosting Nystr\"{o}m algorithm, and the number of learners for ensemble Nystr\"{o}m (\hz{i.e., $p$ above}). For a fixed number of iterations or learners, both the boosting and ensemble Nystr\"{o}m methods utilize the same number of total columns of the kernel matrix, allowing \hz{a fair} comparison of the methods.

\hz{Both plots (a) and (b) in Figure \ref{fig:sim} show that} most of the boosting Nystr\"{o}m \zl{methods} of the form URB-clustering and RRB-clustering outperform the ensemble Nystr\"{o}m method with ridge regression weights (which has the best performance among all the ensemble Nystr\"{o}m methods \cite{kumar2009ensemble}) with the same or even much smaller number of ``weak" learners (standard Nystr\"{o}m approximations). Figure \ref{fig:simb} \hz{also suggests} that even the boosting Nystr\"{o}m methods with exponential \pp{weights} in the strong approximation step achieve smaller relative errors than ensemble Nystr\"{o}m with ridge regression weight method.

\begin{figure}[h!]
  \centering
  \includegraphics[width=0.7\textwidth]{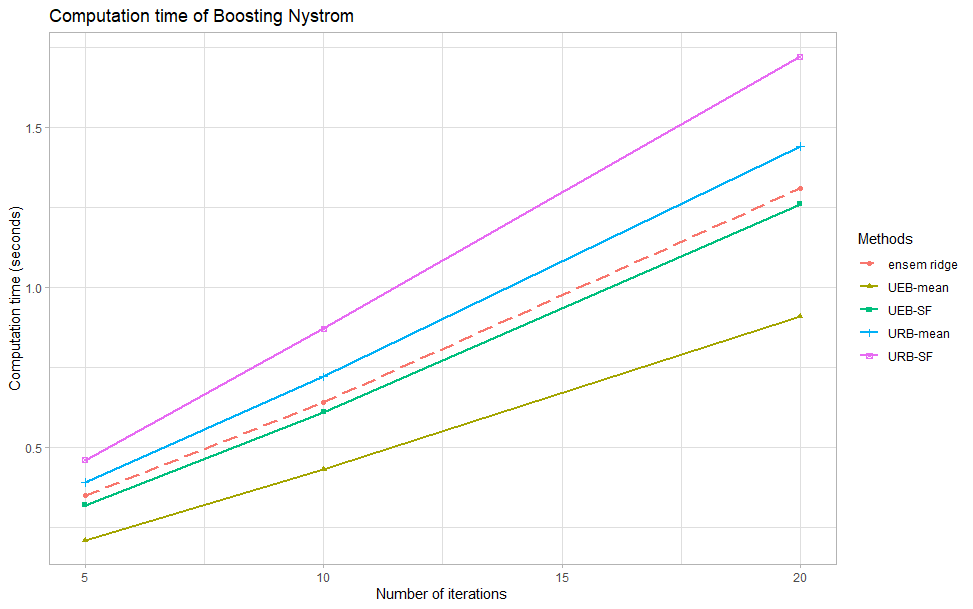}
  \caption{Computational time of some boosting Nystr\"{o}m methods on synthetic Gaussian kernel data.}\label{fig:comp}
\end{figure}

From Figure \ref{fig:sim}, one can also see that \emph{k}-medoids clustering performs better than \emph{k}-means clustering when the same average weighting method is used. The \emph{k}-medoids-SF algorithm performs better than $k$-means but worse than $k$-medoids as one would expect, but Figure \ref{fig:comp} shows that the computational time is significantly reduced when using the $k$-medoids-SF approximation \kh{instead of $k$-medoids.}

To assess how the methods compare to each other \hz{rigorously, we also} performed one-sided, two-sample $t$-tests for all \hz{of the error results} obtained by the ensemble Nystr\"{o}m with the ridge regression weight method. \hz{All of the} $p$-values were less than 0.01, implying that all variations of boosting Nystr\"{o}m \zl{methods} listed in Figure \ref{fig:sim} are significantly better than any ensemble or standard Nystr\"{o}m method.

\hz{To compare the computational time, Figure \ref{fig:comp} shows} the average runtime \hz{for} a given number of iterations of boosting Nystr\"{o}m compared with a single ensemble Nystr\"{o}m approximation with ridge regression weights for the same number of learners. The boosting algorithms which use ridge regression in the strong approximation step take more time than a single ensemble Nystr\"{o}m trial as one would expect, but the time increase is not \kh{dramatic, and the overall complexity appears the same.}  Recall that Figure \ref{fig:sim} shows that the URB variants make up for in accuracy what they lack in speed. However, one can easily see that when exponential weights are used, boosting Nystr\"{o}m gains substantial speed over ridge regression ensemble Nystr\"{o}m.  Note that the UUB variants of boosting Nystr\"{o}m \kh{are} faster than all \hz{the} other methods, but their accuracy does not necessarily beat that of ensemble Nystr\"{o}m. We will explore this further in the subsequent experiments.

From these synthetic experiments, we see that while not all boosting Nystr\"{o}m variants are faster than ensemble Nystr\"{o}m with ridge regression weights, most of them are more accurate, while some are significantly faster. Additionally, we see that many boosting Nystr\"{o}m \zl{methods}  \hz{can outperform} ensemble Nystr\"{o}m with many fewer learners, which significantly speeds up the computation time.

\subsection{Real-world data examples}

In this section, we report the experimental results to show the improvement in the performance of our boosting Nystr\"{o}m \zl{method} \hz{over other competing algorithms}. \hz{We consider five} \zl{real-world data sets}, \hz{whose
sizes and dimensions are listed} in Table \ref{TAB:Data}.  

\begin{table}[ht]
\caption{A summary of the datasets.}\label{TAB:Data}
\centering
\begin{tabular}{l|ccc}
  \hline
 Datasets & {number of points} & dimension \\
  \hline
MNIST~\cite{lecun1998mnist} & 70,000 & 784  \\
{GloVe 6B}~\cite{pennington2014glove} & {400,000} & 300  \\
Covertype~\cite{blackard1998comparison, covtype} & 581,012 & 54 \\
IJCNN~\cite{prokhorov2001ijcnn} & 49,990 & 22  \\
Pendigit~\cite{alimoglu1996methods,pendigits} & 10,992 & 16\\
\hline
\end{tabular}
\end{table}

\begin{figure}[h!]
  \centering
  \begin{subfigure}[b]{0.45\linewidth}
    \includegraphics[width=\linewidth]{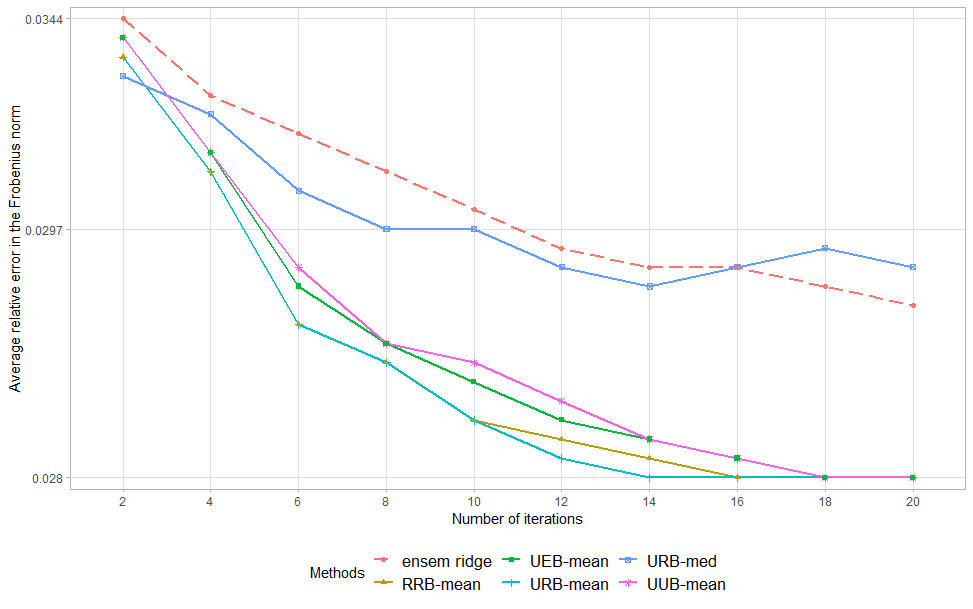}
  \end{subfigure}
  \begin{subfigure}[b]{0.45\linewidth}
    \includegraphics[width=\linewidth]{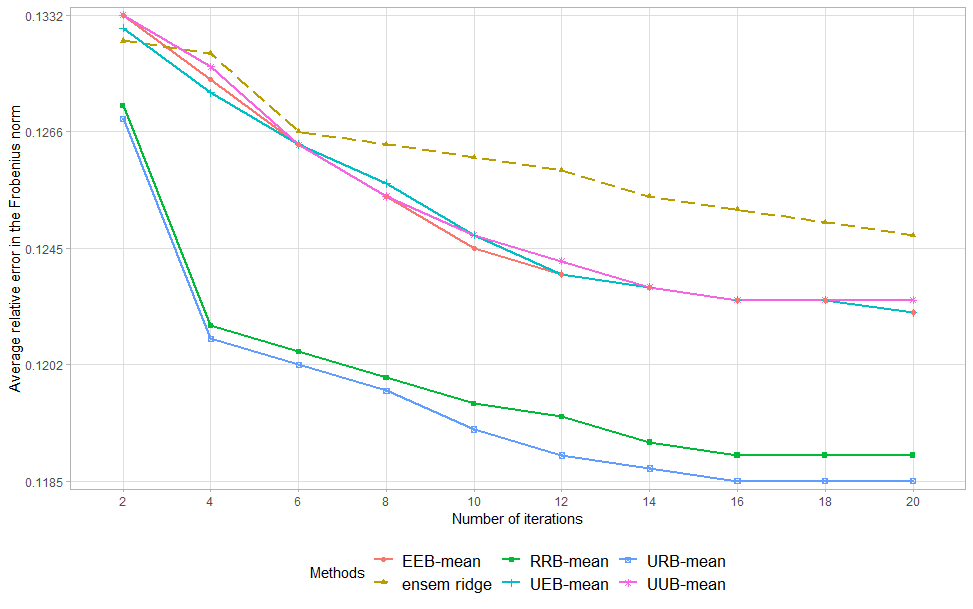}
  \end{subfigure}
  \caption{Average relative errors for 100 replicates of each random algorithm for the MNIST data set with linear kernel (left) and Gaussian kernel with $\sigma=5$ (right) matrices.}
  \label{fig:MNIST}
\end{figure}

\begin{figure}[h!]
  \centering
  \begin{subfigure}[b]{0.45\linewidth}
    \includegraphics[width=\linewidth]{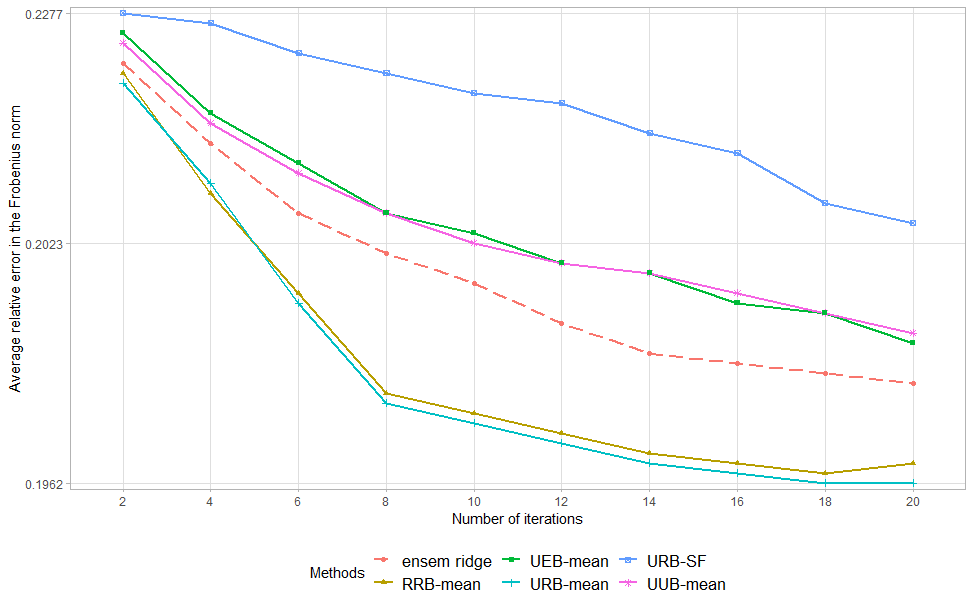}
  \end{subfigure}
  \begin{subfigure}[b]{0.45\linewidth}
    \includegraphics[width=\linewidth]{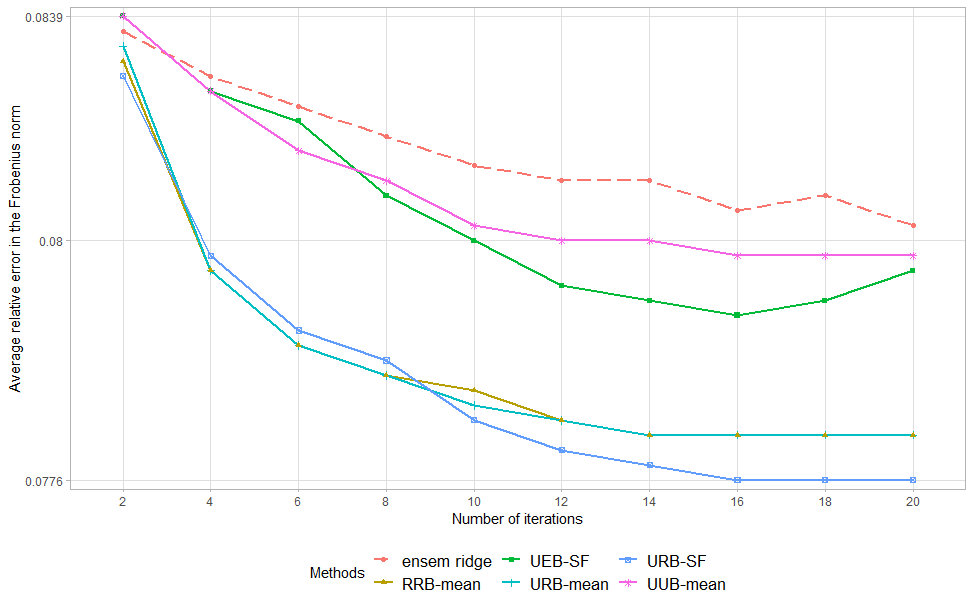}
  \end{subfigure}
  \caption{Average relative errors for 100 replicates of each random algorithm for the GloVe 6B with linear kernel (left) and Gaussian kernel with $\sigma=5$ (right) matrices.}
  \label{fig:NLP}
\end{figure}

\begin{figure}[h!]
  \centering
  \begin{subfigure}[b]{0.45\linewidth}
    \includegraphics[width=\linewidth]{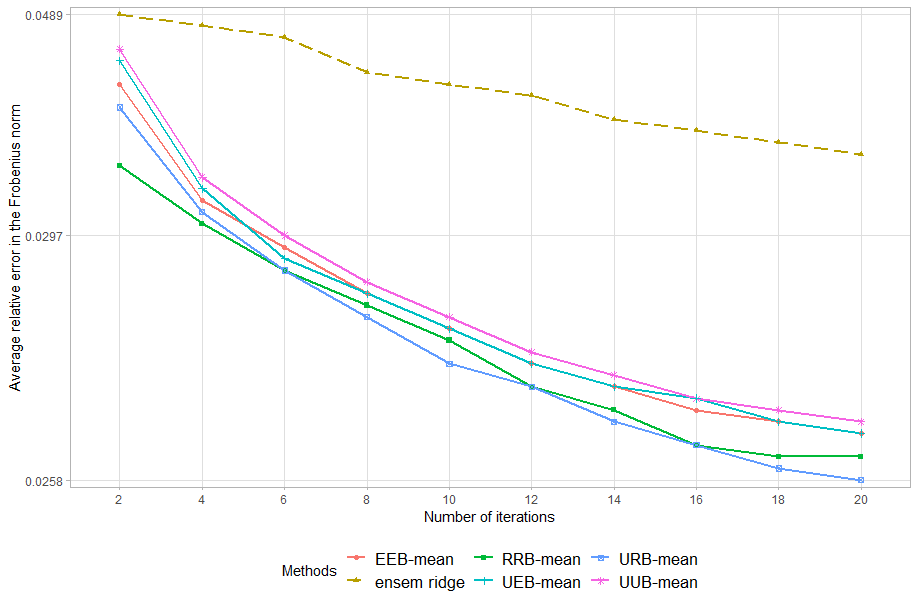}
  \end{subfigure}
  \begin{subfigure}[b]{0.45\linewidth}
    \includegraphics[width=\linewidth]{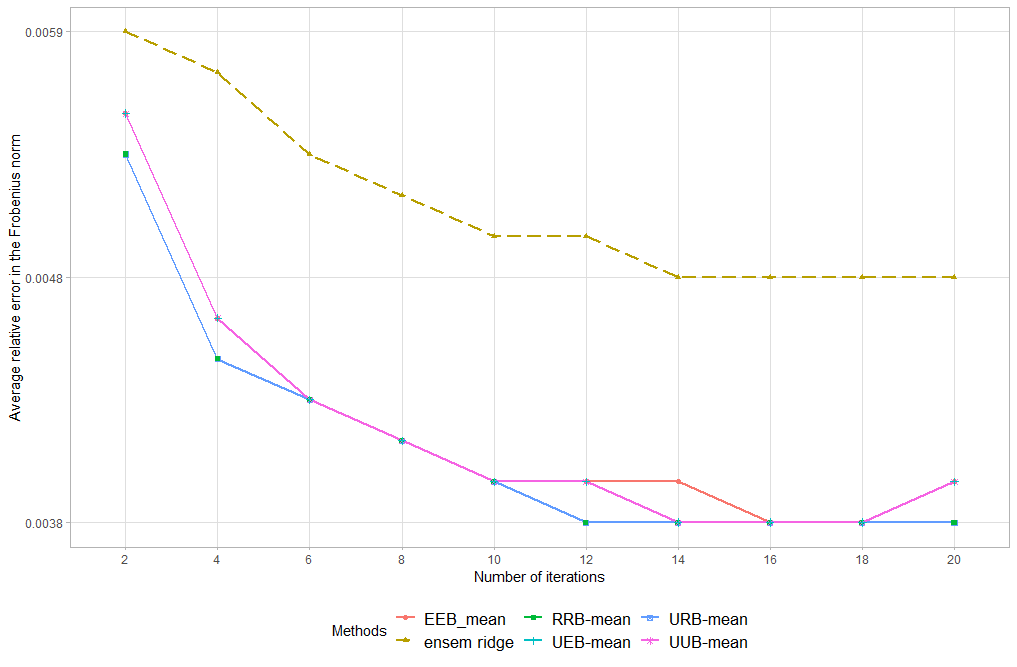}
  \end{subfigure}
    \begin{subfigure}[b]{0.45\linewidth}
    \includegraphics[width=\linewidth]{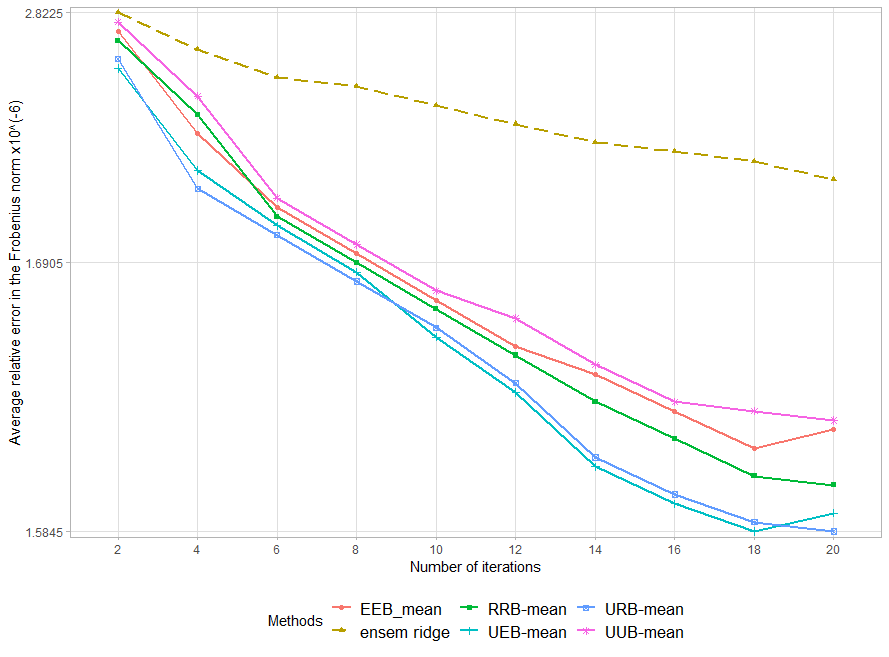}
  \end{subfigure}
  \caption{Average relative errors of boosting Nystr\"{o}m algorithms for Covertype (top left corner), Pendigits (top right corner), and IJCNN (bottom) with Gaussian kernel ($\sigma=5$) matrices.}
  \label{fig:three}
\end{figure}

\hz{For each data set, we randomly sample a subset of 4000 data points to} compute the kernel matrices. \meka{Considering those datasets where features have different scales, we apply column standardization to make the features comparable to each other.} The experimental results shown here are mainly based on the Gaussian kernel with $\sigma = 5$, but we also provide the results of the linear kernel on the first two real data sets. In these experiments, the number of columns for each learner is $m = 120$ and the target rank is set to \hz{be} $k = 50$. The number of columns in the validation sets $V$, $V_1$, and $V_2$ are 600, 20, and 20, respectively. The mean values of relative errors are presented in \zl{Figure \ref{fig:MNIST} (MNIST), Figure \ref{fig:NLP} (GloVe 6B), and Figure \ref{fig:three} (Covertype, IJCNN, and Pendigit)}. 

In Figure \ref{fig:MNIST}, we see that the performance of $k$-means clustering is better than $k$-medoids and $k$-medoids-SF (not shown) for MNIST. 
The algorithms shown in the figure with $k$-means clustering achieve smaller relative errors \kh{for the same number of learners, but note that boosting Nystr\"{o}m achieves the same error rate with half the learners for this data set even for uniform mixture weights.} 
We find that the boosting Nystr\"{o}m \zl{methods} with ridge regression weights outperform the boosting Nystr\"{o}m with exponential \hz{weights} \zl{in general} when \hz{both} use $k$-means clustering, \zl{but one notable exception will be discussed later} \hz{for the} IJCNN data set. Although exponential weights and ridge regression weights are typically better \hz{than uniform weighting since} they are data-dependent, we see that applying them \hz{to} both the boosting step and strong approximation step may not \hz{always} result in the best results. One can see from Figure \ref{fig:MNIST} that URB-mean performs better than RRB-mean throughout the iterative process for the MNIST data set with the Gaussian kernel, and \hz{also} for the linear kernel. The former applies the ridge regression weight method only in the strong approximation step, while the latter applies it in both the boosting step and \hz{the} strong approximation step. One possible explanation for this is that MNIST is incoherent, which is known to be a good case for uniform sampling for Nystr\"{o}m approximations \cite{talwalkar2010matrix}, and so more sophisticated methods are not required. 

Interestingly, for the linear kernel on MNIST \zl{and the Gaussian kernels on Covertype, IJCNN, and Pendigit data sets}, UUB-mean performs similarly to the best methods, but this does not occur \kh{for} \zl{the GloVe 6B dataset or for} the Gaussian kernel \zl{on MNIST}. Again, this is likely due to the known success of uniform sampling for incoherent kernel matrices. \hz{Since} this phenomenon is not true for \kh{all} \hz{other} data sets and kernel matrices (see Figure \ref{fig:NLP} example), one should consider using other variants of boosting Nystr\"{o}m in general.

Figure \ref{fig:NLP} depicts the relative error for approximation of the \zl{GloVe 6B} data set with both linear and Gaussian kernels. By comparing these results with those of Figure \ref{fig:MNIST}, one can see that both the underlying data structure and the kernel function have an impact on the performance of various boosting Nystr\"{o}m \zl{methods}. All the boosting Nystr\"{o}m listed on the graph, with different weighting methods and clustering methods, outperform the ensemble Nystr\"{o}m method with ridge regression weights for the \zl{GloVe 6B} with \hz{the} Gaussian kernel. \hz{In the linear kernel case}, only the boosting Nystr\"{o}m \zl{methods} with ridge regression weights and $k$-means clustering achieve better results than the ensemble Nystr\"{o}m method. One of the benefits of \hz{the proposed} boosting Nystr\"{o}m \hz{framework is that offers a variety of configurations and} options for \hz{a user to explore and make suitable choices for different types} of data sets and kernel functions.

As shown in Figure \ref{fig:three}, \kh{all of the boosting Nystr\"{o}m approximations achieve smaller relative errors than ensemble Nystr\"{o}m, and they also require significantly fewer iterations to achieve the same relative accuracy.} 
The relative errors of our approximations are \kh{very small} for the IJCNN data set because the kernel matrix of IJCNN is \kh{approximately low rank, and our choice of $m$ is larger than the approximate rank.} In addition, we \hz{observe}
\zl{that UEB-mean performs better than RRB-mean, which only occurs} \kh{for} \zl{the Gaussian kernel on IJCNN and \kh{is not observed in other experiments}. The results of boosting Nystr\"{o}m \zl{methods} for Pendigits with the Gaussian kernel} \hz{suggest} that it is hard to achieve higher accuracy after 16 iterations, \hz{yet} boosting Nystr\"{o}m enables us to find the least number of learners for the best performance \hz{with
computational cost reduced significantly.} 

Finally, we \hz{would like to} highlight one of the boosting Nystr\"{o}m \zl{methods}: URB-mean. From all the figures, one can see that this method consistently outperforms ensemble Nystr\"{o}m \kh{and other boosting Nystr\"{o}m methods} for a variety of data sets and kernel matrices, \hz{and} Figure \ref{fig:comp} shows that its computation time is only slightly more than ensemble Nystr\"{o}m with ridge regression weights.  \hz{Therefore,} we conclude that in many settings, URB-mean can be used \hz{to provide high-qualify} Nystr\"{o}m approximation.


\section{Conclusion}\label{SEC:Conclu}

In this paper, we proposed the boosting Nystr\"{o}m framework, which combines the advantages of boosting and those of ensemble Nystr\"{o}m approximations \hz{to estimate} kernel matrices via low-rank approximation. The boosting Nystr\"{o}m \hz{trains a sequence of standard Nystr\"{o}m approximations (weak learners) adaptively followed by weighted aggregation, which is shown to} produce better accuracy than existing Nystr\"o{m} methods for different types of data sets and kernel matrices. The presented  boosting Nystr\"{o}m is a general framework, and, by offering various choices of weights and clustering methods, it is expected to have a wide range of applications.

Experimental results of simulated \hz{experiments and real data analysis} show that our methods gain consistent performance improvement \hz{and only require a smaller} number of Nystr\"{o}m approximations than the ensemble Nystr\"{o}m method with ridge regression weight, which achieves the best performance across all ensemble Nystr\"{o}m methods. Using fewer Nystr\"{o}m approximations allows us to obtain accurate, low-rank approximations of kernel matrices more efficiently. Finally, we found that one variant of boosting Nystr\"{o}m, URB-mean, performs consistently better than ensemble Nystr\"{o}m and most other boosting variants on diverse data and kernel choices. 





\bibliographystyle{plain}

\end{document}